\documentclass[lettersize,journal]{IEEEtran}
\usepackage[T1]{fontenc}
\usepackage{graphicx}
\usepackage{amsmath}
\usepackage{booktabs}
\usepackage{multirow}
\usepackage{cite}
\usepackage{url}
\newcounter{haralgorithm}

\title{OntoAR-HAR: Ontology-Guided Agentic RAG with Personalized Memory for Human Activity Recognition}
\author{Xiangqing Lu, Tao Zhu,~\IEEEmembership{Member,~IEEE}, Liang Yu,
Shuangjian Li, Lulu Ban, Furong Duan, Mingxing Nie,
Liming Chen,~\IEEEmembership{Senior Member,~IEEE}, and Yaping Wan%
\thanks{\emph{(Corresponding author: Tao Zhu.)}}%
\thanks{Xiangqing Lu, Tao Zhu, Liang Yu, Mingxing Nie, and Yaping Wan are with the School of Computer Science, University of South China, Hengyang, China (e-mail: xqlu@stu.usc.edu.cn; tzhu@usc.edu.cn; lyu@stu.usc.edu.cn; niemx@usc.edu.cn; ypwan@usc.edu.cn).}%
\thanks{Shuangjian Li and Liming Chen are with the School of Computer Science and Technology, Dalian University of Technology, Dalian, China (e-mail: shuangjianli@mail.dlut.edu.cn; limingchen0922@dlut.edu.cn).}%
\thanks{Lulu Ban is with Hunan University, Changsha, China (e-mail: llban@hnu.edu.cn).}%
\thanks{Furong Duan is with the School of Computing, Ulster University, Belfast, United Kingdom (e-mail: duan-f@ulster.ac.uk).}}

\begin{document}
\bstctlcite{refstyle:etal}
\maketitle
\begin{abstract}
LLMs increasingly support sensor-based human activity recognition (HAR) by linking motion signals with activity semantics and physical knowledge. Retrieval-augmented generation (RAG) grounds decisions in sensor examples and domain features, enabling training-free target-user inference with less dependence on raw numerical arrays. However, existing LLM/RAG-based HAR typically searches single-level activity spaces using static population knowledge. Consequently, they compare many unrelated candidates using the same static population evidence and overlook personalized motion patterns. We propose OntoAR-HAR, an ontology-guided agentic RAG framework combining static population and personalized knowledge. Its tree-structured activity ontology decomposes classification into local decisions among semantically related activities. The static store supplies node-constrained examples, while offline-discovered executable signal features provide numerical evidence for local distinctions. During inference, the personalized store collects selected user observations, which the LLM integrates with static knowledge and signal features while traversing the ontology to an activity label. Cross-subject and heterogeneous-sensing experiments show complementary gains from hierarchical retrieval and personalized memory across multiple LLM backbones, achieving state-of-the-art performance. By explicitly modeling hierarchical relations across activity granularities, OntoAR-HAR also addresses the long-overlooked semantic inclusion problem and avoids treating related labels as independent categories.
\end{abstract}

\begin{IEEEkeywords}
Human activity recognition, inertial sensors, retrieval-augmented generation, ontologies, personalization.
\end{IEEEkeywords}

\section{Introduction}

Sensor-based human activity recognition (HAR) maps multichannel motion windows to activity labels for applications such as health monitoring, sports analytics, and mobile interaction. Recent LLM-based HAR has progressed from direct prompting toward retrieval-grounded inference. HARGPT studies whether an LLM can select activities from activity descriptions and sensor sequences through direct prompting~\cite{hargpt}. Following the broader RAG paradigm~\cite{rag}, RAG-HAR retrieves sensor observations and statistical descriptors as contextual evidence~\cite{raghar}, while ZARA combines retrieved motion examples with domain knowledge in an LLM-based agent~\cite{zara}. These approaches demonstrate the main advantage of RAG for HAR: observed motion patterns and signal knowledge can be placed directly in the decision context, reducing reliance on an LLM's interpretation of raw numerical arrays alone. Here, RAG denotes retrieval-augmented LLM inference whose final output is an activity label. Its effectiveness depends not only on embedding similarity, but also on how activity candidates and retrieved evidence are organized.

Current RAG-style HAR systems commonly organize activity labels and their examples in a single-level retrieval space. This is problematic when labels express semantic inclusion or share a physical concept. For example, upstairs and downstairs are both forms of walking; sitting, standing, and lying are stationary activities; and eating soup and eating chips share an eating motion but differ in local wrist dynamics. Retrieving across all labels at once introduces candidates that are irrelevant to the current distinction and makes it difficult to associate a physical feature with the local decision where it is useful. Moreover, a static population store contains only source-user sensor examples and is not supplemented with observations from the current user, so it cannot represent that user's characteristic motion amplitude, rhythm, or execution style.

We propose \textbf{OntoAR-HAR}, an ontology-guided agentic RAG framework for HAR that combines a static population knowledge store with a personalized knowledge store. The ontology represents inclusion and sibling relations among activity labels and turns one global retrieval problem into a sequence of local decisions. Static population retrieval evidence and executable physical features are attached to each decision node. At inference time, the separate personalized store records selected observations from the current user. Together, the two stores provide similar sensor examples retrieved from population data and the current user's previous observations. An LLM integrates these examples with numerical feature reports and selects the next ontology branch until it reaches an activity leaf. The framework uses a labeled source collection, a pretrained sensor encoder, and offline feature construction; target inference does not update model parameters.

The contribution of this paper is a structured agentic RAG framework that combines an activity ontology with a personalized knowledge store for HAR. Ontology-guided retrieval limits each reasoning step to a semantically coherent candidate group, addressing activity-label inclusion and local ambiguity, while personalized retrieval complements static population activity knowledge with user-specific observations. We evaluate this design on eight public HAR datasets through cross-subject recognition, component ablations, and heterogeneous sensing conditions.

\section{Related Work}

\paragraph{Representation learning for HAR.}
Convolutional--recurrent models such as DeepConvLSTM~\cite{DeepConvLSTM} established end-to-end sensor representation learning. Self-supervised methods learn transferable temporal structure through contrastive or reconstruction objectives~\cite{cpc,LIMU-BERT,ts2vec,mantis}. Cross-domain HAR further uses adversarial alignment, language supervision, and large-scale motion pretraining~\cite{dann,goat,lanhar,unimts}. These models provide strong learned embeddings; OntoAR-HAR uses such an encoder for retrieval and focuses on how retrieved evidence is organized and combined with activity semantics.

\begin{figure*}[!t]
  \centering
  \includegraphics[width=\textwidth]{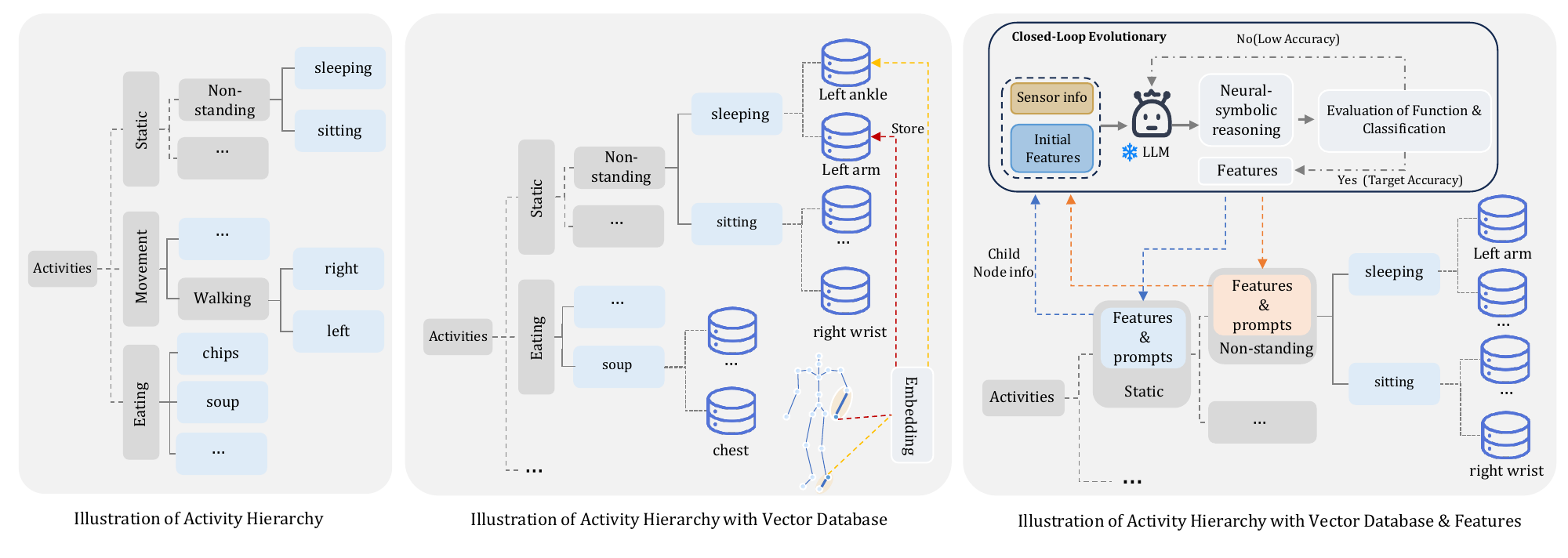}
  \caption{Offline construction. The activity hierarchy is manually grouped and refined using retrieval-only confusions on balanced training samples. Source embeddings form the static population store; an LLM iteratively proposes signal features whose executable implementations are evaluated on confused examples.}
  \label{fig:offline}
\end{figure*}

\begin{table*}[!t]
\caption{Representative Signal-Feature Families Considered during Offline Construction. The Rule-Based Generator Implements Supported Candidates; the Validated Feature Set Is Specific to Each Decision Node.}
\label{tab:feature-families}
\centering
\small
\setlength{\tabcolsep}{4pt}
\renewcommand{\arraystretch}{1.04}
\begin{tabular}{@{}p{0.22\textwidth}p{\dimexpr0.78\textwidth-2\tabcolsep\relax}@{}}
\toprule
Feature family & Representative statistics or derived quantities\\
\midrule
Time-domain statistics & Mean, standard deviation, variance, maximum, minimum, median, root mean square, peak, range, interquartile range, skewness, kurtosis, and signal magnitude area.\\
Frequency-domain statistics & Low-, mid-, and high-frequency band powers and their ratios; dominant frequency and power; secondary peak frequency; spectral centroid, entropy, skewness, and kurtosis; weighted mean frequency; and spectral energy.\\
Short-time Fourier transform & Maximum, mean, and standard deviation in each frequency band, plus statistics of short-time spectral entropy and spectral centroid.\\
Change and impact & Mean, root mean square, and standard deviation of first differences; zero-crossing rate; impact rate of change; and peaks.\\
Periodicity and dependence & First autocorrelation peak, first minimum, and first zero crossing; fourth-order autoregressive coefficients; and recurrence rate.\\
Multiscale complexity & Wavelet-packet energy and permutation entropy.\\
Posture and inter-axis relations & Gravity-direction statistics and inter-axis Pearson correlation coefficients.\\
\bottomrule
\end{tabular}
\end{table*}

The progression from channel-aware supervised models to broadly pretrained encoders is relevant to retrieval quality. Attend and Discriminate weights informative sensor channels when learning an activity representation~\cite{attend}; self-supervised learning has also been studied at the scale of hundreds of thousands of person-days of wearable recordings~\cite{biobankssl}. UniMTS uses unified motion pretraining~\cite{unimts}, while One Model to Fit Them All uses language-assisted cross-dataset representations~\cite{one}. MobHAR instead studies source-free transfer to mobile devices~\cite{mobhar}. These methods address the representation or transfer stage. For a retrieval-grounded decision, a useful embedding is necessary but does not specify which neighboring labels should be compared, which physical feature resolves their ambiguity, or how previous observations from the current person enter the comparison. Our method keeps the sensor encoder fixed and addresses these decisions in the retrieval and reasoning stages.

\paragraph{Language and retrieval for HAR.}
HARGPT studies direct prompting with activity descriptions and sensor sequences~\cite{hargpt}. SensorLLM and SensorLM learn interfaces between continuous signals and language models~\cite{sensorllm,SensorLM}. Retrieval-augmented HAR instead supplies similar observations or feature knowledge as context~\cite{raghar,zara}. OntoAR-HAR differs in organizing retrieval by an activity ontology and maintaining separate static population and user-specific stores. Related neuro-symbolic HAR constrains learning through semantic knowledge~\cite{semanticloss}; our framework uses executable signal features as numerical evidence within ontology-guided retrieval.

Other approaches bring motion and language into a common representation before classification. IMU2CLIP aligns inertial signals with language-grounded representations~\cite{imu2clip}, and COMODO transfers information from video to IMU data through cross-modal distillation~\cite{comodo}. ZeroHAR uses sensor context for zero-shot wearable action recognition~\cite{zerohar}. These approaches make semantic information available to a predictor, whereas an evidence-grounded LLM must additionally decide how to use a retrieved example when its nearest label is only a partial semantic match. Flat retrieval presents all retrieved labels together. Our tree expresses the broader activity first and then associates feature reports with the specific sibling distinction under consideration. This matters for labels such as \emph{walking}, \emph{upstairs}, and \emph{downstairs}, where a similarity match at one granularity need not settle the finer decision.

\paragraph{Adaptation and memory.}
Test-time adaptation methods such as TENT and CoTTA update model parameters or statistics on target streams~\cite{tent,cotta}. OntoAR-HAR does not perform such updates. Its personalized knowledge base is an episodic retrieval mechanism: selected observations become additional context for later decisions. We therefore describe this component as \emph{personalization}, not parameter-based test-time adaptation.

This distinction changes the unit of adaptation. Parameter adaptation changes the mapping from a sensor window to an embedding or label; episodic personalization leaves that mapping fixed and changes the reference examples available to the LLM. The latter can reflect a user's recurring movement pattern without altering the static population store. It also introduces a selection problem: a mistaken pseudo-label could become a future reference. For this reason, the personal store is populated only after an evidence check, and population retrieval remains available at every later decision. Our ablation separates the effect of the hierarchy from the additional effect of this memory.

\section{Method}

\subsection{Problem Formulation and Overview}

Let the labeled source collection be $\mathcal{D}_s=\{(x_i,y_i,b_i)\}$, where $x_i$ is a multichannel sensor window, $y_i\in\mathcal{Y}$ is an activity label, and $b_i$ identifies a sensor view or body location. At inference, the system observes an ordered stream $x_1^q,x_2^q,\ldots$ from a held-out user and predicts labels without gradient updates. Activity labels are organized as a rooted tree $\mathcal{T}=(\mathcal{V},\mathcal{E})$. Leaves correspond to $\mathcal{Y}$ and internal nodes represent semantic groups. A decision at node $v$ considers only $\operatorname{child}(v)$.

\begin{figure*}[!t]
  \centering
  \includegraphics[width=\textwidth]{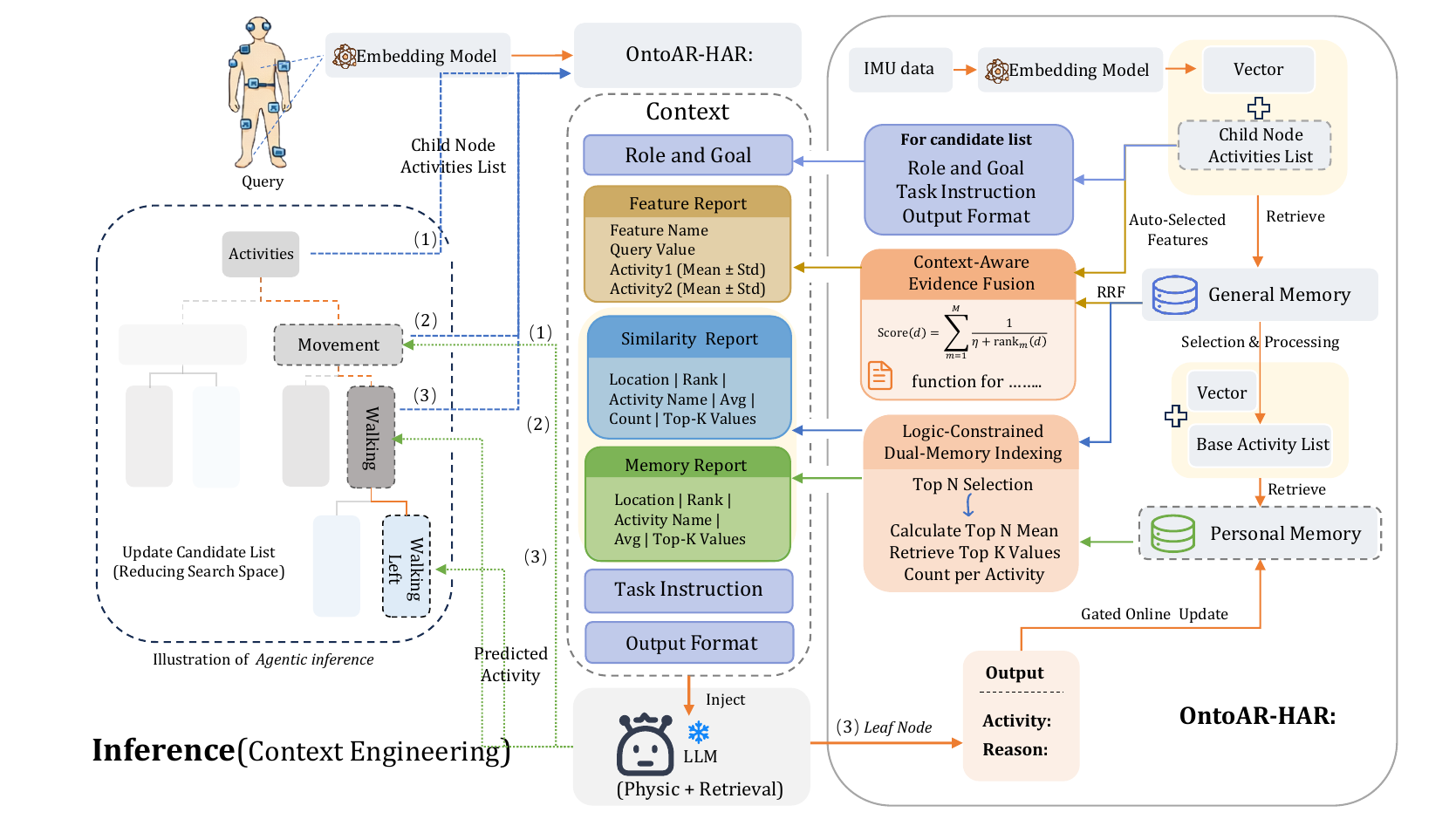}
  \caption{Online inference. At each ontology node, retrieval is restricted to the current subtree. Static population examples, personalized examples, and executable-feature reports form the decision context; the selected child determines the next local problem.}
  \label{fig:inference}
\end{figure*}

OntoAR-HAR has an offline construction phase (Figure~\ref{fig:offline}) and an online inference phase (Figure~\ref{fig:inference}). Offline construction defines the ontology, builds node-constrained vector indexes, and associates executable features with local decisions. Online inference recursively retrieves static population and personalized examples, constructs numerical evidence, and asks an LLM to choose one child at a time.

\subsection{Ontology and Feature Knowledge Construction}

\paragraph{Confusion-guided ontology construction.}
We manually construct the activity hierarchy from retrieval-only predictions on class-balanced activity samples from the training set. We first collect the confusion matrix and place activities with clear confusion relationships into coarse groups, each of which becomes a child of the root. A group may contain several activity labels. We then inspect the confusion submatrix within each group. If a subset is more strongly confused with one another than with the remaining members, we insert an intermediate node above that subset. For example, within a five-activity group, activities 1, 4, and 5 form a nested subgroup when their mutual confusions are notably stronger than their confusions with activities 2 and 3. We repeat this inspection within the resulting subgroups until no further clear subdivision is found. Every activity remains a single leaf, and the completed hierarchy is fixed before evaluation.

The confusion matrix is a design aid rather than an automatically optimized tree objective. Balancing the training examples prevents a frequent class from dominating the observed error pairs. For each proposed internal node, we check that its descendants form a meaningful activity group and that the split gives a local comparison the system can actually make. A class belongs to exactly one branch at a given level; an intermediate node groups leaves without adding a new output label. The resulting tree therefore preserves the original classification task while changing the order in which candidate sets are considered. The hierarchy is built from source training data only and is not revised using the held-out inference users.

\begin{table*}[!t]
\caption{Cross-Subject Results (\%). Acc and Macro-F1 Are Shown in Separate Rows. The Dataset Descriptors Give the Number of Activity Classes and Sensor Channels; OntoAR-HAR Reports the Mean with Standard Deviation as a Subscript over Ten Random Samples. Best Results Are Bold, and Second-Best Results Are Underlined.}
\label{tab:main}
\centering
\footnotesize
\setlength{\tabcolsep}{2.5pt}
\begin{tabular*}{\textwidth}{@{\extracolsep{\fill}}l c *{9}{c}@{}}
\toprule
\multirow{2}{*}{Method} & \multirow{2}{*}{Metric} & \multicolumn{3}{c}{Easy} & \multicolumn{3}{c}{Medium} & \multicolumn{2}{c}{Hard} & \multirow{2}{*}{Avg.}\\
\cmidrule(lr){3-5}\cmidrule(lr){6-8}\cmidrule(lr){9-10}
& & Opp. & UCI-HAR & Shoaib & PAMAP2 & USC-HAD & Mhealth & WISDM & DSADS & \\
\midrule
\# Classes & & 4 & 6 & 7 & 12 & 12 & 12 & 18 & 19 & --\\
\# Channels & & 30 & 6 & 30 & 18 & 6 & 15 & 6 & 30 & --\\
\midrule
\multicolumn{11}{l}{\emph{Cross-modal alignment (full-shot)}}\\
\multirow{2}{*}{ImageBind} & Acc & 82.8 & 90.9 & \textbf{98.5} & 94.3 & 52.1 & 82.9 & 77.2 & 90.0 & 83.6\\
 & F1 & 84.6 & 90.7 & \textbf{98.5} & 94.6 & 52.2 & 83.2 & 76.8 & 89.3 & 83.7\\
\multirow{2}{*}{IMU2CLIP} & Acc & 71.9 & 87.1 & \underline{98.2} & 90.7 & 55.7 & 94.5 & 62.7 & 88.7 & 81.2\\
 & F1 & 70.0 & 87.0 & \underline{98.1} & 91.2 & 52.0 & 95.0 & 62.8 & 88.2 & 80.5\\
\midrule
\multicolumn{11}{l}{\emph{Self-supervised representation learning (full-shot)}}\\
\multirow{2}{*}{TST} & Acc & 89.1 & 91.6 & 73.5 & 88.9 & 66.2 & 86.0 & 70.2 & 80.0 & 80.7\\
 & F1 & 91.1 & 91.4 & 72.3 & 89.8 & 60.9 & 83.8 & 69.1 & 76.1 & 79.3\\
\multirow{2}{*}{TARNet} & Acc & 82.6 & 91.1 & 67.9 & 92.4 & 57.6 & 82.9 & 58.5 & 30.7 & 70.5\\
 & F1 & 83.4 & 91.2 & 66.1 & 90.6 & 48.7 & 77.5 & 58.5 & 23.0 & 67.4\\
\multirow{2}{*}{TS2Vec} & Acc & 80.7 & 92.1 & 87.4 & 91.8 & 56.1 & 93.3 & 69.5 & 84.3 & 81.9\\
 & F1 & 83.3 & \underline{92.1} & 87.2 & 92.9 & 53.6 & 93.8 & 68.8 & 82.3 & 81.8\\
\midrule
\multicolumn{11}{l}{\emph{Supervised and pretrained HAR models (full-shot)}}\\
\multirow{2}{*}{THAT} & Acc & 83.1 & 86.8 & 92.3 & \textbf{96.5} & 53.9 & 89.0 & 60.3 & 82.1 & 80.5\\
 & F1 & 84.5 & 86.7 & 92.2 & \textbf{96.7} & 55.0 & 89.9 & 60.9 & 79.0 & 80.6\\
\multirow{2}{*}{IMUGPT} & Acc & 84.8 & 87.0 & 61.7 & 85.6 & 41.1 & 71.3 & 67.8 & 71.9 & 71.4\\
 & F1 & 84.9 & 87.0 & 61.8 & 83.8 & 42.4 & 63.6 & 67.4 & 71.0 & 70.2\\
\multirow{2}{*}{TimesNet} & Acc & 80.0 & 88.5 & 82.4 & 88.0 & 58.3 & 81.7 & 67.4 & 77.3 & 77.9\\
 & F1 & 82.4 & 88.6 & 82.6 & 88.0 & 48.7 & 79.1 & 67.1 & 78.5 & 76.9\\
\multirow{2}{*}{GPT4TS} & Acc & 83.6 & 88.4 & 73.5 & 92.6 & 61.7 & 87.8 & 63.8 & 79.9 & 78.9\\
 & F1 & 86.1 & 88.5 & 74.1 & 93.4 & 58.9 & 85.2 & 64.5 & 78.0 & 78.6\\
\multirow{2}{*}{SHARE} & Acc & 89.0 & \underline{92.2} & 77.5 & 94.1 & 72.9 & \textbf{97.0} & 77.0 & 92.0 & 86.5\\
 & F1 & 90.2 & \underline{92.1} & 78.0 & \underline{94.8} & 66.7 & \textbf{97.1} & 76.5 & 91.9 & 85.9\\
\multirow{2}{*}{UniMTS} & Acc & 89.1 & 91.5 & \textbf{98.5} & \underline{96.2} & 56.5 & \underline{96.3} & \underline{80.5} & \underline{92.9} & \underline{87.7}\\
 & F1 & 90.1 & 91.7 & \textbf{98.5} & \textbf{96.7} & 61.6 & \underline{96.7} & \underline{79.9} & \underline{92.6} & \textbf{88.5}\\
\midrule
\multicolumn{11}{l}{\emph{Retrieval-based inference}}\\
\multirow{2}{*}{ZARA} & Acc & \textbf{92.5} & 90.0 & 97.1 & 76.7 & 60.0 & 86.3 & 65.6 & 84.2 & 81.6\\
 & F1 & \textbf{92.5} & 90.0 & 97.1 & 76.9 & 60.1 & 86.1 & 64.1 & 84.4 & 81.4\\
\multirow{2}{*}{OntoAR-HAR$_{\mathrm{Gemini\,2.0\,Flash}}$} & Acc & $85.7_{\pm2.4}$ & $90.1_{\pm5.3}$ & $96.8_{\pm2.0}$ & $82.6_{\pm1.5}$ & $\underline{74.8}_{\pm3.2}$ & $74.6_{\pm2.9}$ & $77.9_{\pm1.8}$ & $89.8_{\pm3.4}$ & $84.0_{\pm1.1}$\\
 & F1 & $85.5_{\pm2.4}$ & $90.9_{\pm4.1}$ & $96.8_{\pm2.0}$ & $83.2_{\pm1.6}$ & $\underline{75.2}_{\pm3.3}$ & $74.0_{\pm3.2}$ & $77.2_{\pm1.9}$ & $89.2_{\pm3.5}$ & $84.0_{\pm0.8}$\\
\multirow{2}{*}{OntoAR-HAR$_{\mathrm{Gemini\,2.5\,Flash}}$} & Acc & $89.8_{\pm2.3}$ & $89.6_{\pm5.4}$ & $94.5_{\pm4.3}$ & $84.5_{\pm1.7}$ & $70.7_{\pm2.2}$ & $83.9_{\pm3.0}$ & $77.0_{\pm2.4}$ & $90.9_{\pm2.4}$ & $85.1_{\pm1.7}$\\
 & F1 & $89.9_{\pm2.3}$ & $89.5_{\pm5.4}$ & $93.4_{\pm4.8}$ & $85.2_{\pm2.1}$ & $70.8_{\pm2.2}$ & $83.6_{\pm3.0}$ & $76.2_{\pm2.4}$ & $90.4_{\pm2.7}$ & $84.9_{\pm1.7}$\\
\multirow{2}{*}{OntoAR-HAR$_{\mathrm{Gemini\,3.0\,Flash}}$} & Acc & $\underline{91.3}_{\pm1.6}$ & $\mathbf{93.1}_{\pm5.7}$ & $96.4_{\pm2.8}$ & $86.7_{\pm1.1}$ & $\mathbf{80.2}_{\pm3.8}$ & $84.0_{\pm4.1}$ & $\mathbf{80.7}_{\pm3.0}$ & $\mathbf{93.2}_{\pm3.8}$ & $\mathbf{88.2}_{\pm2.3}$\\
 & F1 & $\underline{91.5}_{\pm1.6}$ & $\mathbf{92.8}_{\pm5.9}$ & $96.5_{\pm2.9}$ & $87.3_{\pm1.1}$ & $\mathbf{80.1}_{\pm3.4}$ & $84.6_{\pm2.6}$ & $\mathbf{80.2}_{\pm3.3}$ & $\mathbf{92.8}_{\pm4.3}$ & $\underline{88.2}_{\pm2.0}$\\
\bottomrule
\end{tabular*}
\end{table*}

\paragraph{Iterative feature construction.}
At each ambiguous node, an LLM proposes a structured list of feature names, physical interpretations, and calculations from the child labels and sensor metadata. Table~\ref{tab:feature-families} summarizes representative candidate families. A rule-based generator implements supported proposals. On examples misclassified by retrieval-only inference, we test execution and measure separation by mean standardized feature distance to retrieved activity references. Errors and remaining misclassifications inform the next prompt. Algorithm~\ref{alg:feature-generation} specifies the loop and retained outputs. Each internal node $v$ stores $\Theta_v=(\mathcal{F}_v,\pi_v)$, where $\mathcal{F}_v$ is the selected executable feature set and $\pi_v$ describes the local decision.

This construction connects a feature to a specific ambiguity. For example, a distinction between stationary postures calls for different evidence from a distinction between level walking and stair motion. Candidate descriptions alone are insufficient: the proposal must be translated into an executable calculation for the available sensor channels. The loop first checks whether that calculation runs on the hard examples, then uses retrieval-only misses to identify where its reported values remain uninformative. Its feedback contains both execution failures and residual decision errors, so a later proposal can revise the feature definition rather than merely restate the activity names. The generated functions and their node-specific descriptions are fixed before target-user inference.

\par\medskip
\refstepcounter{haralgorithm}\label{alg:feature-generation}
\noindent\begin{minipage}{\columnwidth}
\footnotesize
\hrule\vspace{3pt}
\textbf{Algorithm \theharalgorithm: Iterative feature generation at node $v$}\par\vspace{2pt}
\begin{tabular}{@{}r@{\hspace{0.5em}}p{\dimexpr\columnwidth-2.2em\relax}@{}}
  & \textbf{Input:} Child labels, sensor metadata, retrieval-only errors $H$.\\
1 & prompt := InitialPrompt(labels, sensors); history := empty.\\
2 & \textbf{for} round = 1 to 5 \textbf{do}\\
3 & \quad features := ParseJSON(LLM(prompt)); retry if invalid.\\
4 & \quad function := GenerateByRules(features); retry if unloadable.\\
5 & \quad errors := Execute(function, $H$).\\
6 & \quad accuracy, misses := DistanceProxy(function, first 20 of $H$).\\
7 & \quad Record features, function, errors, and accuracy.\\
8 & \quad \textbf{if} no errors and (accuracy exceeds 0.8 or is unavailable), break.\\
9 & \quad prompt := FeedbackPrompt(features, errors, misses).\\
10 & \textbf{end for}; save history and the last generated function.
\end{tabular}
\vspace{3pt}\hrule
\end{minipage}
\par\medskip

\subsection{Static Population and Personalized Knowledge Stores}

For each sensor view, the \emph{static population store} $\mathcal{D}_{g}$ contains source embeddings and labels. A logical mask restricts retrieval at node $v$ to examples whose labels occur below $v$:
\begin{equation}
 \mathcal{R}_{g}(x_q,v)=\operatorname{topk}\{s(e_q,e_i):y_i\in\operatorname{leaf}(v)\},
\end{equation}
where $e_q$ and $e_i$ are encoder embeddings and $s$ is cosine similarity. The indexes are physically shared but logically filtered, avoiding a separate database for every node.

The \emph{personalized store} $\mathcal{D}_{p}$ has the same representation but is initialized empty for each inference user. It contains previously observed query windows paired with accepted predictions. Retrieval first obtains the active labels from $\mathcal{D}_{g}$ and then searches $\mathcal{D}_{p}$ under the same label and node constraints. The two stores remain separate so that static population evidence is always visible and personal examples cannot replace the source reference distribution.

This order makes the static collection an anchor for the current local decision. The source search identifies plausible labels under the current ontology node; a subsequent personal search asks whether accepted windows from this user support any of those labels. When no personal example satisfies the constraints, the decision proceeds with source examples and feature evidence. As the personal store grows, each accepted window remains associated with the user and its predicted leaf, so a later branch search can retrieve it only where that leaf is relevant. This avoids treating a user's history as an unrestricted global neighbor list.

For $M$ sensor views, ranked lists are combined using reciprocal rank fusion (RRF)~\cite{rrf}:
\begin{equation}
 \operatorname{RRF}(d)=\sum_{m=1}^{M}\frac{1}{\eta+\operatorname{rank}_m(d)},
\end{equation}
with $\eta=60$. The retrieved examples define reference distributions for features in $\mathcal{F}_v$. OntoAR-HAR reports the query value, reference mean and standard deviation, and standardized deviation for each child. This makes the LLM compare explicit numerical evidence rather than calculate signal statistics from raw arrays.

Rank fusion is useful here because embeddings from different sensor views may have different score ranges: it combines their ordering without requiring raw similarity scores to be on the same scale. The feature report performs a different task. For a candidate child, it summarizes how the query's executable feature values compare with retrieved examples associated with that child. A large standardized deviation flags a mismatch for inspection, while the population and personal examples retain their separate provenance. Thus, view fusion determines which examples enter the context, and the numerical report helps interpret those examples at the active ontology node.

\subsection{Recursive Inference and Memory Update}

At node $v$, the context $\mathcal{C}_v$ contains the local instruction $\pi_v$, similar examples retrieved and fused from $\mathcal{D}_{g}$ and $\mathcal{D}_{p}$, and the feature report. The LLM returns a child and a short evidence record:
\begin{equation}
 \hat c_v=\arg\max_{c\in\operatorname{child}(v)}P_{\mathrm{LLM}}(c\mid\mathcal{C}_v).
\end{equation}
The agent moves to $\hat c_v$ until reaching a leaf $\hat y$. This coarse-to-fine path directly represents cases such as activity $\rightarrow$ locomotion $\rightarrow$ walking direction, so a feature is evaluated only where it distinguishes siblings.

After prediction, a prompt-based gate decides whether $(x_q,\hat y)$ enters $\mathcal{D}_{p}$. The gate checks agreement between static population retrieval, personal retrieval when available, and the node feature ranges; ambiguous or contradictory cases are not inserted. Subsequent samples from the same user can retrieve accepted episodes, allowing the context to reflect recurring personal motion patterns without changing encoder or LLM weights.

\section{Experiments}

\subsection{Setup}

We use Opportunity~\cite{opp}, UCI-HAR~\cite{ucihar}, Shoaib~\cite{shoaib}, PAMAP2~\cite{pamap}, USC-HAD~\cite{uschad}, Mhealth~\cite{mhealth}, WISDM~\cite{wisdm}, and DSADS~\cite{dsads}. These datasets span 4--19 activities, 6--30 channels, and waist, wrist, chest, ankle, arm, leg, and pocket placements. Database subjects and inference subjects are disjoint. In the cross-subject evaluation, inference subsets are sampled without replacement and balanced by activity and user; their sizes are 200, 240, 210, 240, 240, 240, 288, and 190 windows, respectively. Cross-subject results use seeds 0, 1, 42, 123, 520, 527, 1024, 2026, 2778, and 3407. Accuracy measures leaf-label correctness, while macro-F1 gives equal weight to the per-class F1 scores.

\begin{table}[!t]
\caption{Semantic-Inclusion Evaluation. The Upper Block Reports Dataset-Level Hierarchical F1; the Lower Block Gives Overall Metrics. Percentage Changes Are in Points.}
\label{tab:semantic-inclusion}
\centering
{\small
\setlength{\tabcolsep}{0.5mm}
\begin{tabular}{lrrr@{\hspace{1mm}}lrrr}
\toprule
\multicolumn{8}{c}{Hierarchical F1 (\%)}\\
Dataset & Single & Hier. & $\Delta$ & Dataset & Single & Hier. & $\Delta$\\
\midrule
DSADS & 95.26 & 96.05 & +0.79 & Shoaib & 99.29 & 99.52 & +0.23\\
Mhealth & 89.31 & 91.04 & +1.73 & UCI-HAR & 96.67 & 97.50 & +0.83\\
Opp. & 89.50 & 92.75 & +3.25 & USC-HAD & 78.54 & 81.11 & +2.57\\
PAMAP2 & 89.03 & 89.58 & +0.55 & WISDM & 81.77 & 82.99 & +1.22\\
Avg. & 89.92 & 91.32 & +1.40 & & & &\\
\midrule
Acc. (\%) & 84.24 & 87.17 & +2.92 & Macro-F1 (\%) & 84.36 & 87.10 & +2.75\\
Tree dist. & 0.428 & 0.366 & -0.062 & Cross-br. (\%) & 5.18 & 4.96 & -0.22\\
\bottomrule
\end{tabular}
}
\end{table}

\begin{figure}[t]
  \centering
  \includegraphics[width=\columnwidth]{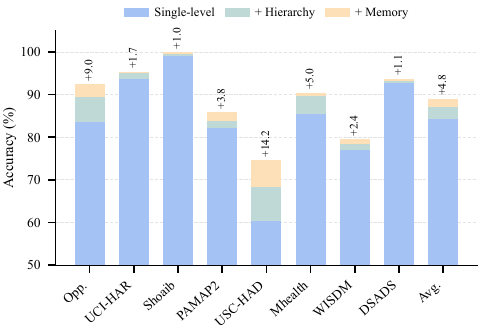}
  \caption{Controlled component study with Gemini 2.5 Flash. Each vertical bar starts from single-level retrieval; the upper segments show the incremental gains from hierarchy and personalized memory.}
  \label{fig:ablation}
\end{figure}

\begin{table*}[t]
\caption{Heterogeneous-Dataset Results (\%) over Six Harmonized Activity Categories without Body-Location Filtering. Acc and Macro-F1 Are Shown Side by Side. Best Results Are Bold, and Second-Best Results Are Underlined.}
\label{tab:cross}
\centering
\resizebox{\textwidth}{!}{%
\begin{tabular}{l*{6}{cc}}
\toprule
Method & \multicolumn{2}{c}{UCI-HAR} & \multicolumn{2}{c}{PAMAP2} & \multicolumn{2}{c}{USC-HAD} & \multicolumn{2}{c}{Mhealth} & \multicolumn{2}{c}{DSADS} & \multicolumn{2}{c}{Average}\\
\cmidrule(lr){2-3}\cmidrule(lr){4-5}\cmidrule(lr){6-7}\cmidrule(lr){8-9}\cmidrule(lr){10-11}\cmidrule(lr){12-13}
& Acc & F1 & Acc & F1 & Acc & F1 & Acc & F1 & Acc & F1 & Acc & F1\\
\midrule
ImageBind & 14.1 & 5.0 & 14.6 & 8.1 & 11.8 & 7.3 & 7.9 & 1.7 & 2.1 & 1.2 & 10.1 & 4.7\\
IMU2CLIP & 17.8 & 11.4 & 1.9 & 1.1 & 12.8 & 9.3 & 7.9 & 1.3 & 4.6 & 1.1 & 9.0 & 4.8\\
IMUGPT & 1.1 & 0.3 & 8.9 & 1.5 & 6.0 & 6.9 & 9.8 & 2.5 & 7.5 & 2.0 & 6.7 & 2.6\\
HARGPT & 15.0 & 12.7 & 11.1 & 2.1 & 9.5 & 3.6 & 10.4 & 7.4 & 5.8 & 3.4 & 10.4 & 5.8\\
ZARA & 16.7 & 4.9 & 17.1 & 8.6 & 16.7 & 4.8 & 20.0 & 6.7 & 16.7 & 6.8 & 17.4 & 6.4\\
UniMTS & 35.2 & 22.0 & 47.2 & 43.6 & 30.5 & 27.8 & \textbf{70.7} & \textbf{61.8} & 31.5 & 23.7 & 43.0 & 35.8\\
\midrule
OntoAR-HAR$_{\mathrm{Gemini\,2.0}}$ & 45.83 & 41.63 & 54.29 & 46.63 & \textbf{41.67} & \textbf{38.57} & 50.00 & 36.71 & \underline{36.67} & 22.66 & 45.69 & 37.24\\
OntoAR-HAR$_{\mathrm{Gemini\,2.5}}$ & \underline{48.33} & \textbf{45.26} & \underline{56.43} & \underline{51.96} & \underline{36.67} & 35.04 & 50.00 & 36.92 & \textbf{41.67} & \textbf{32.37} & \underline{46.62} & \underline{40.31}\\
OntoAR-HAR$_{\mathrm{Gemini\,3.0}}$ & \textbf{49.17} & \underline{43.94} & \textbf{60.00} & \textbf{57.49} & \textbf{41.67} & \underline{37.61} & \underline{51.00} & \underline{39.64} & \textbf{41.67} & \underline{31.87} & \textbf{48.70} & \textbf{42.11}\\
\bottomrule
\end{tabular}}
\end{table*}

\begin{figure*}[t]
  \centering
  \includegraphics[width=0.97\textwidth]{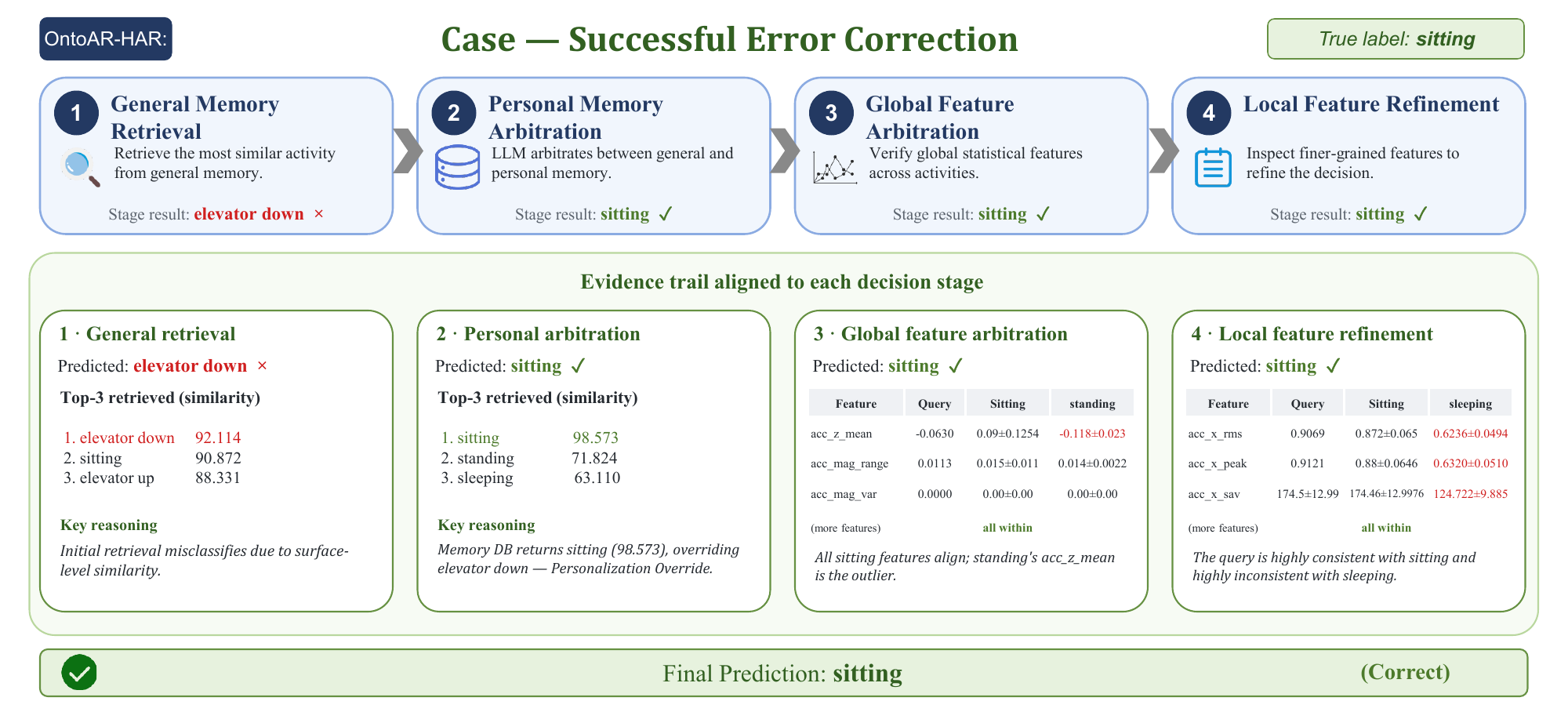}
  \caption{Example decision path. Static population retrieval, personalized retrieval, and node-specific feature reports are kept separate in the context, allowing the final decision to be traced through the ontology.}
  \label{fig:qualitative}
\end{figure*}

The eight datasets vary in more than label count. Their sensors cover different body positions and channel configurations, so nearest-neighbor retrieval must operate over different signal views. The cross-subject setting tests whether source-user examples remain useful for a person excluded from the database. Balancing the inference subset by both activity and user keeps the aggregate from being driven by the most common recorded action or the most heavily sampled participant. The controlled component comparison uses the cross-subject setting and Gemini 2.5 Flash, isolating the effect of the decision structure and memory within that evaluation configuration.

\paragraph{Signal windows and subject separation.}
We retain the segmentation used for each dataset rather than forcing every source into the same number of time steps. Opportunity uses non-overlapping 2-s windows of 60 samples; UCI-HAR uses its 2.56-s, 128-sample windows with 50\% overlap. Shoaib and Mhealth use non-overlapping 2-s windows of 100 samples, while PAMAP2 and USC-HAD use 2-s windows of 200 samples. WISDM uses 10-s windows of 200 samples after timestamp alignment of smartwatch accelerometer and gyroscope recordings; DSADS supplies predefined 5-s windows of 125 samples. These choices explain the different channel/window sizes reported in Table~\ref{tab:efficiency}.

The source and inference partitions are disjoint by subject. Opportunity reserves subject 4; Shoaib reserves subjects 1 and 9; PAMAP2 reserves 5 and 6; USC-HAD reserves 13 and 14; Mhealth reserves 1 and 6; and DSADS reserves 2 and 4. UCI-HAR retains its predefined nine-subject test partition. For WISDM, eight of the 47 users with aligned smartwatch recordings are set aside for inference. The stated inference sizes result from class-balanced sampling within these held-out partitions, with the available examples per user also accounted for. This separation keeps a user's earlier accepted query windows distinct from the source examples used to initialize the static store.

Baselines cover cross-modal models (ImageBind~\cite{imagebind} and IMU2CLIP~\cite{imu2clip}), self-supervised temporal models (TST~\cite{TST}, TARNet~\cite{tarnet}, and TS2Vec~\cite{ts2vec}), supervised or pretrained HAR models (THAT~\cite{that}, IMUGPT~\cite{IMUGPT2.0}, TimesNet~\cite{timesnet}, GPT4TS~\cite{gpt4ts}, SHARE~\cite{share}, and UniMTS~\cite{unimts}), and retrieval-based ZARA~\cite{zara}. These methods use different supervision regimes, so Table~\ref{tab:main} labels full-shot and retrieval settings rather than treating them as identical. OntoAR-HAR uses Mantis embeddings and temperature zero for LLM calls. Cross-view feature-reference fusion uses the standard RRF constant $\eta=60$ and retains at most 100 fused items; these settings are fixed across datasets and LLM backbones rather than tuned per dataset. Experiments run on Windows 11 with an Intel Core i5-12600KF CPU, 32\,GB RAM, and an NVIDIA GeForce RTX 4060 Ti GPU (8\,GB), using Python 3.9, PyTorch 2.8, NumPy 1.26.4, and scikit-learn 1.6.1; LLM inference uses hosted APIs.

\paragraph{Baseline implementation.}
The cross-modal full-shot baselines use a downstream classifier with their sensor embeddings. ImageBind starts from the pretrained \texttt{imagebind\_huge} checkpoint and tunes the backbone and linear head; IMU2CLIP is first pretrained on Ego4D because official pretrained weights are unavailable, then its encoder and classifier are tuned for the target task. Among the temporal baselines, TST is pretrained before downstream fine-tuning, TARNet jointly optimizes reconstruction and classification, and TS2Vec learns contrastive representations before a dataset-specific linear readout. The direct-prompt HARGPT input is downsampled to 10 Hz to fit its prompt setting. For the zero-shot setting, IMUGPT uses a DeepConvLSTM backbone without its supervised distribution-calibration phase. ZARA follows its default pairwise feature knowledge setup with temperature zero. These details explain why the supervised and retrieval rows of Table~\ref{tab:main} are informative comparisons but do not share an identical training budget.

\begin{table*}[t]
\caption{Detailed Performance of Different LLM Backbones in the Controlled Comparison (\%). Accuracy and Macro-F1 Are Shown Side by Side. Best Results Are Bold, and Second-Best Results Are Underlined.}
\label{tab:backbones}
\centering
\setlength{\tabcolsep}{2.5pt}
\resizebox{\textwidth}{!}{%
\begin{tabular}{l*{9}{cc}}
\toprule
Backbone & \multicolumn{2}{c}{Opp.} & \multicolumn{2}{c}{UCI} & \multicolumn{2}{c}{Shoaib} & \multicolumn{2}{c}{PAMAP2} & \multicolumn{2}{c}{USC} & \multicolumn{2}{c}{Mhealth} & \multicolumn{2}{c}{WISDM} & \multicolumn{2}{c}{DSADS} & \multicolumn{2}{c}{Avg.}\\
\cmidrule(lr){2-3}\cmidrule(lr){4-5}\cmidrule(lr){6-7}\cmidrule(lr){8-9}\cmidrule(lr){10-11}\cmidrule(lr){12-13}\cmidrule(lr){14-15}\cmidrule(lr){16-17}\cmidrule(lr){18-19}
& Acc & F1 & Acc & F1 & Acc & F1 & Acc & F1 & Acc & F1 & Acc & F1 & Acc & F1 & Acc & F1 & Acc & F1\\
\midrule
Gemini 2.0 Flash & 83.00 & 82.71 & 92.50 & 92.33 & 96.19 & 96.19 & 85.42 & 85.94 & 71.67 & 71.60 & 85.83 & 85.51 & 78.12 & 77.30 & 90.53 & 90.44 & 85.41 & 85.25\\
Gemini 2.5 Flash & \textbf{92.50} & \textbf{92.51} & \textbf{95.42} & \textbf{95.38} & \textbf{100.00} & \textbf{100.00} & 85.83 & 86.01 & 74.58 & 74.60 & \textbf{90.42} & \underline{89.33} & \underline{79.51} & 78.54 & \underline{93.68} & 93.05 & \underline{88.99} & 88.68\\
Gemini 3.0 Flash & \underline{90.50} & \underline{90.62} & \underline{94.17} & 94.09 & \textbf{100.00} & \textbf{100.00} & \textbf{89.17} & \underline{89.49} & \textbf{82.50} & \textbf{82.01} & \underline{89.58} & \textbf{89.45} & \textbf{80.90} & \textbf{80.27} & \textbf{94.74} & \textbf{94.79} & \textbf{90.20} & \textbf{90.09}\\
DeepSeek R1 & 80.50 & 80.28 & 94.12 & \underline{94.17} & 96.67 & 96.66 & 85.00 & 85.22 & 71.25 & 71.50 & 78.33 & 77.80 & \underline{79.51} & \underline{78.66} & 92.63 & 91.98 & 84.75 & 84.53\\
Grok 4 & 85.50 & 85.41 & 92.50 & 92.44 & \underline{99.52} & \underline{99.52} & \underline{87.50} & 87.90 & 72.50 & 72.41 & 85.00 & 84.58 & 78.82 & 77.83 & 93.15 & 92.78 & 86.81 & 86.61\\
GPT-5 & 89.50 & 89.65 & 92.50 & 92.42 & \textbf{100.00} & \textbf{100.00} & \textbf{89.17} & \textbf{89.56} & \underline{77.50} & \underline{77.43} & 89.17 & 88.97 & 78.47 & 77.33 & \textbf{94.74} & \underline{94.54} & 88.88 & \underline{88.74}\\
\bottomrule
\end{tabular}}
\end{table*}

\begin{table}[t]
\caption{Per-Query Efficiency with Gemini 2.5 Flash. C/W Denotes Channels/Window; Ret. Is Local Retrieval Time; Tokens (K) Are Prompt/Completion/Total, with the Total Computed before Rounding; Cost Is in USD.}
\label{tab:efficiency}
\centering
\setlength{\tabcolsep}{2pt}
\resizebox{\columnwidth}{!}{%
\begin{tabular}{lrrrr}
\toprule
Dataset (C/W) & Ret. (s) & LLM (s) & Tokens (K) & Cost\\
\midrule
DSADS (30/125) & 0.057 & 1.39 & 1.91/0.35/2.25 & 0.001442\\
Mhealth (15/100) & 0.036 & 1.67 & 1.29/0.31/1.60 & 0.001170\\
Opportunity (30/60) & 0.077 & 2.59 & 2.71/0.50/3.20 & 0.002055\\
PAMAP2 (18/200) & 0.054 & 2.52 & 2.31/0.45/2.76 & 0.001822\\
Shoaib (30/100) & 0.092 & 1.48 & 1.61/0.47/2.08 & 0.001655\\
UCI-HAR (6/128) & 0.019 & 4.96 & 1.03/0.18/1.22 & 0.000768\\
USC-HAD (6/200) & 0.022 & 3.99 & 1.33/0.22/1.55 & 0.000945\\
WISDM (6/200) & 0.041 & 4.72 & 1.00/0.26/1.25 & 0.000947\\
\bottomrule
\end{tabular}}
\vspace{-6pt}
\end{table}

\subsection{Cross-Subject Results and Component Analysis}

Table~\ref{tab:main} reports 88.2 average accuracy and F1 for OntoAR-HAR with Gemini 3.0 Flash, versus 87.7/88.5 for UniMTS and 81.6/81.4 for ZARA. UniMTS is a demanding but non-equivalent reference because it uses full-shot optimization, whereas OntoAR-HAR performs retrieval-based inference without task-specific or target-user parameter updates; ZARA is the closer protocol match. OntoAR-HAR is strongest on USC-HAD (80.2 accuracy versus ZARA's 60.0) and competitive on WISDM and DSADS, while UniMTS remains stronger on PAMAP2 and Mhealth. ZARA slightly leads on Opportunity and Shoaib, consistent with hierarchy being most useful when several activities share a broader motion concept.

Figure~\ref{fig:ablation} isolates the two structural components with Gemini 2.5 Flash. Average accuracy rises from 84.0 with single-level retrieval to 87.2 with hierarchy and 89.0 after adding personalized memory. The hierarchy helps most on closely related subclasses, while the additional memory gain shows that current-user examples complement static population evidence.

The two increments answer different questions. The first retains retrieval and LLM reasoning but replaces the global candidate set with successive sibling comparisons, testing whether label organization improves the use of otherwise similar evidence. The second retains that hierarchy and permits accepted prior windows from the inference user to appear beside the population neighbors. Their cumulative result is consistent with complementary roles: the tree selects the comparison to make, and memory changes the examples available for making it. The experiment does not require parameter updates to either the sensor encoder or the LLM.

\paragraph{Semantic-inclusion evaluation.}
We compare controlled single-level and hierarchy outputs using path-aware metrics. Hierarchical F1 measures overlap between the true and predicted ontology paths after removing the common root; tree distance is their shortest-path length; cross-branch error counts predictions outside the true top-level branch. Table~\ref{tab:semantic-inclusion} shows gains on all datasets: average Hierarchical F1 rises from 89.92 to 91.32, while tree distance falls from 0.428 to 0.366 and cross-branch error from 5.18\% to 4.96\%. Together with the 84.24-to-87.17 accuracy gain, hierarchy improves both leaf recognition and the semantic proximity of residual errors.

\subsection{Heterogeneous-Dataset Evaluation}

We form a mixed benchmark from UCI-HAR, PAMAP2, USC-HAD, Mhealth, and DSADS using six harmonized activity categories and no body-location filtering. Database and inference users remain disjoint, and retrieval searches the heterogeneous collection globally.

The six evaluation activities are standing, sitting, lying, walking, upstairs, and downstairs. Labels with common or closely corresponding meanings are mapped to this output space, but the underlying windows come from different sensor placements and collection protocols. Keeping those placements rather than filtering to a single location makes the task a joint test of activity semantics and sensing heterogeneity. In particular, a label match across datasets does not imply that its raw channel values or nearest-neighbor geometry match. The ontology supplies a common sequence of local label decisions, while the embedding and retrieved examples still have to carry the signal-level evidence for each query.

Table~\ref{tab:cross} shows 48.70 accuracy and 42.11 macro-F1 for OntoAR-HAR with Gemini 3.0, versus 43.0/35.8 for UniMTS. OntoAR-HAR leads on four of five datasets, indicating that local semantic decisions transfer across mixed sensing configurations; Mhealth also shows that representation quality and source coverage still affect retrieved evidence.

\subsection{Backbone and Reasoning Effects}

Table~\ref{tab:backbones} is a fixed-subset backbone comparison, not a replacement for Table~\ref{tab:main}'s ten-sample estimates. Gemini-family average Acc/F1 increases from 85.41/85.25 to 88.99/88.68 and 90.20/90.09; DeepSeek R1, Grok 4, and GPT-5 confirm compatibility but also backbone sensitivity.

\begin{figure}[t]
  \centering
  \includegraphics[width=\columnwidth]{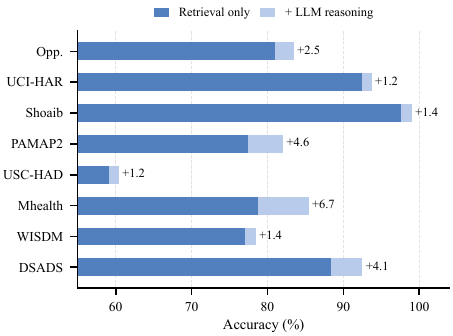}
  \caption{Retrieval-only accuracy and the change obtained after LLM reasoning with the same retrieved evidence. Positive changes occur on all eight datasets.}
  \label{fig:llm-reasoning}
\end{figure}

Figure~\ref{fig:llm-reasoning} isolates the Gemini 2.5 Flash decision stage. Mean accuracy rises from 81.51 to 84.40, showing a contribution beyond similarity aggregation, although this test does not separate general reasoning from model-specific evidence integration.

This comparison is especially relevant to the feature construction stage. The retrieved list is not only a vote over stored labels: it supplies class-conditioned examples against which executable signal features can be evaluated. The LLM receives the resulting values and local candidate descriptions, then selects a branch. The positive change across the eight datasets indicates that this additional decision step uses information beyond the retrieval-only label ranking. Because the backbone is fixed in this comparison, the observed difference belongs to the evidence-to-label decision stage rather than a change in the underlying sensor encoder.

\subsection{Inference Cost and Reasoning Example}

Table~\ref{tab:efficiency} reports Gemini 2.5 Flash inference cost. Retrieval takes 0.019--0.092 seconds per query, versus 1.39--4.96 seconds for each LLM decision; prompts contain 1.00K--2.71K tokens and cost USD 0.000768--0.002055. Sequential LLM calls therefore dominate latency, while the ontology limits each call to the current node's children.

Figure~\ref{fig:qualitative} illustrates the evidence flow. Static retrieval initially favors \emph{elevator down} (92.114) over \emph{sitting} (90.872), but personalized retrieval supplies a previously accepted \emph{sitting} episode (98.573). Subsequent global and local feature comparisons separate \emph{sitting} from \emph{standing} and \emph{sleeping}. The final path is thus supported by inspectable population examples, user-specific episodes, and numerical feature checks.

The case shows why nearest-neighbor similarity need not be the final activity decision. The static store makes an elevator-related label plausible, but the personal example provides stronger evidence for this user's sitting pattern. A feature comparison then checks that interpretation against other stationary labels. Each source of information has a visible role: the static examples prevent the personalized match from being considered in isolation, the personal record captures a recurring user pattern, and the node features address a narrower motion distinction. This interpretation is limited to the illustrated query; the dataset-level evidence for the components comes from the ablation and cross-subject tables.

\subsection{Discussion}

The ontology and personalized store act at complementary levels: the former restricts each comparison through label relations, while the latter changes its evidence using current-user observations without replacing the source reference. Executable features add signal-level evidence where needed. The LLM therefore integrates bounded candidates, retrieved examples, and feature statistics rather than raw arrays or an unconstrained label list, leaving an inspectable decision path.

\section{Conclusion and Future Work}

OntoAR-HAR combines an activity ontology with static population and personalized knowledge stores for sensor-based HAR. The ontology represents inclusion relations among labels and restricts retrieval and feature reasoning to local sibling decisions. Personalized episodes complement static population evidence without parameter updates. Across eight datasets, the hierarchy and memory provide additive gains, while the heterogeneous-dataset experiment indicates that the same structure can help under mixed sensing conditions. The framework thus connects label-level semantics, source-domain sensor evidence, and observations from the current user in a single recursive inference process.

Future work will automate and scale ontology construction beyond the current manual, confusion-guided procedure, including the generation and validation of intermediate concepts for larger activity vocabularies. We will also develop more complete memory gates that model confidence and long-term drift, and improve feature selection with broader libraries, automated validation, and robust fallback mechanisms. These extensions will reduce manual ontology design and make personalization more reliable over longer user streams.

\bibliographystyle{IEEEtran}
\bibliography{OntoAR}

@IEEEtranBSTCTL{refstyle:etal,
  CTLuse_forced_etal = "yes",
  CTLmax_names_forced_etal = "6",
  CTLnames_show_etal = "1"
}

@String{Computing = "Computing" }

@String{Computer = "{IEEE} Computer" }

@article{attend,
  title = {Attend and {{Discriminate}}: {{Beyond}} the {{State-of-the-Art}} for {{Human Activity Recognition Using Wearable Sensors}}},
  shorttitle = {Attend and {{Discriminate}}},
  author = {Abedin, Alireza and Ehsanpour, Mahsa and Shi, Qinfeng and Rezatofighi, Hamid and Ranasinghe, Damith C.},
  year = 2021,
  month = mar,
  journal = {Proceedings of the ACM on Interactive, Mobile, Wearable and Ubiquitous Technologies},
  volume = {5},
  number = {1},
  pages = {1--22},
  issn = {2474-9567},
  doi = {10.1145/3448083},
  urldate = {2026-01-20},
  langid = {american},
  note={doi: 10.1145/3448083}
}

@article{dsads,
  title = {Comparative Study on Classifying Human Activities with Miniature Inertial and Magnetic Sensors},
  author = {Altun, Kerem and Barshan, Billur and Tun{\c c}el, Orkun},
  year = 2010,
  month = oct,
  journal = {Pattern Recognition},
  volume = {43},
  number = {10},
  pages = {3605--3620},
  issn = {00313203},
  doi = {10.1016/j.patcog.2010.04.019},
  urldate = {2026-01-20},
  copyright = {https://www.elsevier.com/tdm/userlicense/1.0/},
  langid = {american},
  note={doi: 10.1016/j.patcog.2010.04.019}
}

@inproceedings{ucihar,
  title={A Public Domain Dataset for Human Activity Recognition Using Smartphones},
  author={Anguita, Davide and Ghio, Alessandro and Oneto, Luca and Parra, Xavier and Reyes-Ortiz, Jorge Luis},
  booktitle={Proc. European Symposium on Artificial Neural Networks, Computational Intelligence and Machine Learning (ESANN)},
  pages={437--442},
  year={2013},
  url={https://www.esann.org/sites/default/files/proceedings/legacy/es2013-84.pdf}
}

@article{mhealth,
  title = {Design, Implementation and Validation of a Novel Open Framework for Agile Development of Mobile Health Applications},
  author = {Banos, Oresti and Villalonga, Claudia and Garcia, Rafael and Saez, Alejandro and Damas, Miguel and {Holgado-Terriza}, Juan A and Lee, Sungyong and Pomares, Hector and Rojas, Ignacio},
  year = 2015,
  journal = {BioMedical Engineering OnLine},
  volume = {14},
  number = {Suppl 2},
  pages = {S6},
  issn = {1475-925X},
  doi = {10.1186/1475-925X-14-S2-S6},
  urldate = {2026-01-20},
  langid = {american},
  note={doi: 10.1186/1475-925X-14-S2-S6}
}

@inproceedings{rrf,
  title = {Reciprocal Rank Fusion Outperforms Condorcet and Individual Rank Learning Methods},
  booktitle = {Proceedings of the 32nd International {{ACM SIGIR}} Conference on {{Research}} and Development in Information Retrieval},
  author = {Cormack, Gordon V. and Clarke, Charles L A and Buettcher, Stefan},
  year = 2009,
  month = jul,
  series = {{{SIGIR}} '09},
  pages = {758--759},
  publisher = {Association for Computing Machinery},
  address = {New York, NY, USA},
  doi = {10.1145/1571941.1572114},
  urldate = {2026-01-20},
  isbn = {978-1-60558-483-6},
  langid = {american},
  note={doi: 10.1145/1571941.1572114}
}

@inproceedings{mantis,
  title={{Mantis}: Lightweight Foundation Model for Time Series Classification},
  author={Vasilii Feofanov and Songkang Wen and Shifeng Xie and Simon Roschmann and Marius Alonso and Hongbo Guo and Romain Ilbert and Malik Tiomoko and Quentin Bouniot and Zeynep Akata and Lujia Pan and Jianfeng Zhang and Ievgen Redko},
  booktitle={Forty-third International Conference on Machine Learning},
  year={2026},
  url={https://openreview.net/forum?id=gbJMAjXLZ4}
}

@article{dann,
  title={Domain-adversarial training of neural networks},
  author={Ganin, Yaroslav and Ustinova, Evgeniya and Ajakan, Hana and Germain, Pascal and Larochelle, Hugo and Laviolette, Fran{\c{c}}ois and March, Mario and Lempitsky, Victor},
  journal={Journal of Machine Learning Research},
  volume={17},
  number={59},
  pages={1--35},
  year={2016}
}

@inproceedings{imagebind,
  title={Imagebind: One embedding space to bind them all},
  author={Girdhar, Rohit and El-Nouby, Alaaeldin and Liu, Zhuang and Singh, Mannat and Alwala, Kalyan Vasudev and Joulin, Armand and Misra, Ishan},
  booktitle={Proceedings of the IEEE/CVF conference on computer vision and pattern recognition},
  pages={15180--15190},
  year={2023},
  doi={10.1109/CVPR52729.2023.01457},
  note={doi: 10.1109/CVPR52729.2023.01457}
}

@inproceedings{hargpt,
  title={Hargpt: Are llms zero-shot human activity recognizers?},
  author={Ji, Sijie and Zheng, Xinzhe and Wu, Chenshu},
  booktitle={2024 IEEE International Workshop on Foundation Models for Cyber-Physical Systems \& Internet of Things (FMSys)},
  pages={38--43},
  year={2024},
  organization={IEEE},
  doi={10.1109/FMSys62467.2024.00011},
  note={doi: 10.1109/FMSys62467.2024.00011}
}

@article{IMUGPT2.0,
  title = {{{IMUGPT}} 2.0: {{Language-Based Cross Modality Transfer}} for {{Sensor-Based Human Activity Recognition}}},
  shorttitle = {{{IMUGPT}} 2.0},
  author = {Leng, Zikang and Bhattacharjee, Amitrajit and Rajasekhar, Hrudhai and Zhang, Lizhe and Bruda, Elizabeth and Kwon, Hyeokhyen and Pl{\"o}tz, Thomas},
  year = 2024,
  month = aug,
  journal = {Proceedings of the ACM on Interactive, Mobile, Wearable and Ubiquitous Technologies},
  volume = {8},
  number = {3},
  pages = {1--32},
  issn = {2474-9567},
  doi = {10.1145/3678545},
  urldate = {2025-03-10},
  langid = {american},
  note={doi: 10.1145/3678545}
}

@inproceedings{sensorllm,
  title={Sensorllm: Aligning large language models with motion sensors for human activity recognition},
  author={Li, Zechen and Deldari, Shohreh and Chen, Linyao and Xue, Hao and Salim, Flora D},
  booktitle={Proceedings of the 2025 Conference on Empirical Methods in Natural Language Processing},
  pages={354--379},
  year={2025},
  doi={10.18653/v1/2025.emnlp-main.19},
  note={doi: 10.18653/v1/2025.emnlp-main.19}
}

@inproceedings{zara,
  title = {Zara: Training-free motion time-series reasoning via evidence-grounded {LLM} agents},
  author = {Li, Zechen and Chen, Baiyu and Xue, Hao and Salim, Flora D},
  booktitle = {Proceedings of the 64th Annual Meeting of the Association for Computational Linguistics (Volume 1: Long Papers)},
  pages = {14986--15008},
  year = {2026},
  doi={10.18653/v1/2026.acl-long.684},
  note={doi: 10.18653/v1/2026.acl-long.684}
}

@article{goat,
  title = {{{GOAT}}: {{A Generalized Cross-Dataset Activity Recognition Framework}} with {{Natural Language Supervision}}},
  shorttitle = {{{GOAT}}},
  author = {Miao, Shenghuan and Chen, Ling},
  year = 2024,
  month = nov,
  journal = {Proceedings of the ACM on Interactive, Mobile, Wearable and Ubiquitous Technologies},
  volume = {8},
  number = {4},
  pages = {1--28},
  issn = {2474-9567},
  doi = {10.1145/3699736},
  urldate = {2026-01-20},
  langid = {american},
  note={doi: 10.1145/3699736}
}

@inproceedings{imu2clip,
  title = {{{IMU2CLIP}}: {{Language-grounded Motion Sensor Translation}} with {{Multimodal Contrastive Learning}}},
  shorttitle = {{{IMU2CLIP}}},
  booktitle = {Findings of the {{Association}} for {{Computational Linguistics}}: {{EMNLP}} 2023},
  author = {Moon, Seungwhan and Madotto, Andrea and Lin, Zhaojiang and Saraf, Aparajita and Bearman, Amy and Damavandi, Babak},
  year = 2023,
  pages = {13246--13253},
  publisher = {Association for Computational Linguistics},
  address = {Singapore},
  doi = {10.18653/v1/2023.findings-emnlp.883},
  urldate = {2026-01-20},
  langid = {american},
  note={doi: 10.18653/v1/2023.findings-emnlp.883}
}

@misc{cpc,
  title = {Representation {{Learning}} with {{Contrastive Predictive Coding}}},
  author = {van den Oord, Aaron and Li, Yazhe and Vinyals, Oriol},
  year = 2018,
  number = {arXiv:1807.03748},
  eprint = {1807.03748},
  primaryclass = {cs},
  publisher = {arXiv},
  doi = {10.48550/arXiv.1807.03748},
  urldate = {2026-01-20},
  archiveprefix = {arXiv},
  langid = {american},
  note={doi: 10.48550/arXiv.1807.03748}
}

@article{DeepConvLSTM,
  title = {Deep {{Convolutional}} and {{LSTM Recurrent Neural Networks}} for {{Multimodal Wearable Activity Recognition}}},
  author = {Ord{\'o}{\~n}ez, Francisco and Roggen, Daniel},
  year = 2016,
  month = jan,
  journal = {Sensors},
  volume = {16},
  number = {1},
  pages = {115},
  issn = {1424-8220},
  doi = {10.3390/s16010115},
  urldate = {2026-01-20},
  langid = {american},
  note={doi: 10.3390/s16010115}
}

@inproceedings{pamap,
  title = {Introducing a {{New Benchmarked Dataset}} for {{Activity Monitoring}}},
  booktitle = {2012 16th {{International Symposium}} on {{Wearable Computers}}},
  author = {Reiss, Attila and Stricker, Didier},
  year = 2012,
  month = jun,
  pages = {108--109},
  publisher = {IEEE},
  address = {Newcastle, United Kingdom},
  doi = {10.1109/ISWC.2012.13},
  urldate = {2026-01-20},
  isbn = {978-0-7695-4697-1 978-1-4673-1583-8},
  langid = {american},
  note={doi: 10.1109/ISWC.2012.13}
}

@inproceedings{opp,
  title = {Collecting Complex Activity Datasets in Highly Rich Networked Sensor Environments},
  booktitle = {2010 {{Seventh International Conference}} on {{Networked Sensing Systems}} ({{INSS}})},
  author = {Roggen, Daniel and Calatroni, Alberto and Rossi, Mirco and Holleczek, Thomas and Forster, Kilian and Troster, Gerhard and Lukowicz, Paul and Bannach, David and Pirkl, Gerald and Ferscha, Alois and Doppler, Jakob and Holzmann, Clemens and Kurz, Marc and Holl, Gerald and Chavarriaga, Ricardo and Sagha, Hesam and Bayati, Hamidreza and Creatura, Marco and Millan, Jose Del R.},
  year = 2010,
  month = jun,
  pages = {233--240},
  publisher = {IEEE},
  address = {Kassel, Germany},
  doi = {10.1109/INSS.2010.5573462},
  urldate = {2026-01-20},
  isbn = {978-1-4244-7911-5},
  langid = {american},
  note={doi: 10.1109/INSS.2010.5573462}
}

@article{shoaib,
  title = {Fusion of {{Smartphone Motion Sensors}} for {{Physical Activity Recognition}}},
  author = {Shoaib, Muhammad and Bosch, Stephan and Incel, Ozlem and Scholten, Hans and Havinga, Paul},
  year = 2014,
  month = jun,
  journal = {Sensors},
  volume = {14},
  number = {6},
  pages = {10146--10176},
  issn = {1424-8220},
  doi = {10.3390/s140610146},
  urldate = {2026-01-20},
  langid = {american},
  note={doi: 10.3390/s140610146}
}

@article{one,
  title = {One {{Model}} to {{Fit Them All}}: {{Universal IMU-based Human Activity Recognition}} with {{LLM-assisted Cross-dataset Representation}}},
  shorttitle = {One {{Model}} to {{Fit Them All}}},
  author = {Wei, Qingxin and Huang, Jiaming and Gao, Yi and Dong, Wei},
  year = 2025,
  month = sep,
  journal = {Proceedings of the ACM on Interactive, Mobile, Wearable and Ubiquitous Technologies},
  volume = {9},
  number = {3},
  pages = {1--22},
  issn = {2474-9567},
  doi = {10.1145/3749509},
  urldate = {2026-01-20},
  langid = {american},
  note={doi: 10.1145/3749509}
}

@misc{wisdm,
  title={{WISDM} Smartphone and Smartwatch Activity and Biometrics Dataset},
  author={Weiss, Gary M.},
  howpublished={UCI Machine Learning Repository, Dataset},
  year={2019},
  doi={10.24432/C5HK59},
  note={doi: 10.24432/C5HK59}
}

@article{mobhar,
  title = {{{MobHAR}}: {{Source-free Knowledge Transfer}} for {{Human Activity Recognition}} on {{Mobile Devices}}},
  shorttitle = {{{MobHAR}}},
  author = {Xue, Meng and Zhu, Yinan and Xie, Wentao and Wang, Zhixian and Chen, Yanjiao and Jiang, Kui and Zhang, Qian},
  year = 2025,
  month = mar,
  journal = {Proceedings of the ACM on Interactive, Mobile, Wearable and Ubiquitous Technologies},
  volume = {9},
  number = {1},
  pages = {1--24},
  issn = {2474-9567},
  doi = {10.1145/3712620},
  urldate = {2026-01-20},
  langid = {american},
  note={doi: 10.1145/3712620}
}

@inproceedings{LIMU-BERT,
  title = {{{LIMU-BERT}}: {{Unleashing}} the {{Potential}} of {{Unlabeled Data}} for {{IMU Sensing Applications}}},
  shorttitle = {{{LIMU-BERT}}},
  booktitle = {Proceedings of the 19th {{ACM Conference}} on {{Embedded Networked Sensor Systems}}},
  author = {Xu, Huatao and Zhou, Pengfei and Tan, Rui and Li, Mo and Shen, Guobin},
  year = 2021,
  month = nov,
  series = {{{SenSys}} '21},
  pages = {220--233},
  publisher = {Association for Computing Machinery},
  address = {New York, NY, USA},
  doi = {10.1145/3485730.3485937},
  urldate = {2026-01-20},
  isbn = {978-1-4503-9097-2},
  langid = {american},
  note={doi: 10.1145/3485730.3485937}
}

@article{lanhar,
  title={Large Language Model-guided Semantic Alignment for Human Activity Recognition},
  author={Yan, Hua and Tan, Heng and Ding, Yi and Zhou, Pengfei and Namboodiri, Vinod and Yang, Yu},
  journal={Proceedings of the ACM on Interactive, Mobile, Wearable and Ubiquitous Technologies},
  volume={9},
  number={4},
  pages={1--25},
  year={2025},
  publisher={ACM New York, NY, USA},
  doi={10.1145/3770652},
  note={doi: 10.1145/3770652}
}

@inproceedings{SensorLM,
  title={{SensorLM}: Learning the Language of Wearable Sensors},
  author={Zhang, Yuwei and Ayush, Kumar and Qiao, Siyuan and Heydari, A. Ali and Narayanswamy, Girish and Xu, Max and Metwally, Ahmed and Xu, Jinhua and Garrison, Jake and Xu, Xuhai and Althoff, Tim and Liu, Yun and Kohli, Pushmeet and Zhan, Jiening and Malhotra, Mark and Patel, Shwetak and Mascolo, Cecilia and Liu, Xin and McDuff, Daniel and Yang, Yuzhe},
  booktitle={Advances in Neural Information Processing Systems},
  volume={38},
  pages={52274--52310},
  year={2025},
  doi={10.52202/085713-1559},
  note={doi: 10.52202/085713-1559}
}

@article{unimts,
  title={UniMTS: Unified Pre-training for Motion Time Series},
  author={Zhang, Xiyuan and Teng, Diyan and Chowdhury, Ranak Roy and Li, Shuheng and Hong, Dezhi and Gupta, Rajesh and Shang, Jingbo},
  journal={Advances in Neural Information Processing Systems},
  volume={37},
  pages={107469--107493},
  year={2024},
  doi={10.52202/079017-3414},
  note={doi: 10.52202/079017-3414}
}

@inproceedings{uschad,
  title = {{{USC-HAD}}: A Daily Activity Dataset for Ubiquitous Activity Recognition Using Wearable Sensors},
  shorttitle = {{{USC-HAD}}},
  booktitle = {Proceedings of the 2012 {{ACM Conference}} on {{Ubiquitous Computing}}},
  author = {Zhang, Mi and Sawchuk, Alexander A.},
  year = 2012,
  month = sep,
  series = {{{UbiComp}} '12},
  pages = {1036--1043},
  publisher = {Association for Computing Machinery},
  address = {New York, NY, USA},
  doi = {10.1145/2370216.2370438},
  urldate = {2026-01-20},
  isbn = {978-1-4503-1224-0},
  langid = {american},
  note={doi: 10.1145/2370216.2370438}
}

@article{rag,
  title={Retrieval-augmented generation for knowledge-intensive nlp tasks},
  author={Lewis, Patrick and Perez, Ethan and Piktus, Aleksandra and Petroni, Fabio and Karpukhin, Vladimir and Goyal, Naman and K{\"u}ttler, Heinrich and Lewis, Mike and Yih, Wen-tau and Rockt{\"a}schel, Tim and others},
  journal={Advances in Neural Information Processing Systems},
  volume={33},
  pages={9459--9474},
  year={2020}
}

@article{comodo,
  title={{COMODO}: Cross-Modal Video-to-{IMU} Distillation for Efficient Egocentric Human Activity Recognition},
  author={Chen, Baiyu and Wongso, Wilson and Li, Zechen and Khaokaew, Yonchanok and Xue, Hao and Salim, Flora D.},
  journal={Proc. ACM Interact. Mob. Wearable Ubiquitous Technol.},
  volume={10},
  number={2},
  pages={1--29},
  year={2026},
  doi={10.1145/3810218},
  note={doi: 10.1145/3810218}
}

@inproceedings{zerohar,
  title={Zerohar: Sensor context augments zero-shot wearable action recognition},
  author={Chowdhury, Ranak Roy and Kapila, Ritvik and Panse, Ameya and Zhang, Xiyuan and Teng, Diyan and Kulkarni, Rashmi and Hong, Dezhi and Gupta, Rajesh K and Shang, Jingbo},
  booktitle={Proceedings of the AAAI Conference on Artificial Intelligence},
  volume={39},
  number={15},
  pages={16046--16054},
  year={2025},
  doi={10.1609/aaai.v39i15.33762},
  note={doi: 10.1609/aaai.v39i15.33762}
}

@inproceedings{TST,
  title={A transformer-based framework for multivariate time series representation learning},
  author={Zerveas, George and Jayaraman, Srideepika and Patel, Dhaval and Bhamidipaty, Anuradha and Eickhoff, Carsten},
  booktitle={Proceedings of the 27th ACM SIGKDD Conference on Knowledge Discovery \& Data Mining},
  pages={2114--2124},
  year={2021},
  doi={10.1145/3447548.3467401},
  note={doi: 10.1145/3447548.3467401}
}

@inproceedings{tarnet,
  title={Tarnet: Task-aware reconstruction for time-series transformer},
  author={Chowdhury, Ranak Roy and Zhang, Xiyuan and Shang, Jingbo and Gupta, Rajesh K and Hong, Dezhi},
  booktitle={Proceedings of the 28th ACM SIGKDD conference on knowledge discovery and data Mining},
  pages={212--220},
  year={2022},
  doi={10.1145/3534678.3539329},
  note={doi: 10.1145/3534678.3539329}
}

@inproceedings{ts2vec,
  title={Ts2vec: Towards universal representation of time series},
  author={Yue, Zhihan and Wang, Yujing and Duan, Juanyong and Yang, Tianmeng and Huang, Congrui and Tong, Yunhai and Xu, Bixiong},
  booktitle={Proceedings of the AAAI Conference on Artificial Intelligence},
  volume={36},
  number={8},
  pages={8980--8987},
  year={2022},
  doi={10.1609/aaai.v36i8.20881},
  note={doi: 10.1609/aaai.v36i8.20881}
}

@article{biobankssl,
  title={Self-supervised learning for human activity recognition using 700,000 person-days of wearable data},
  author={Yuan, Hang and Chan, Shing and Creagh, Andrew P and Tong, Catherine and Acquah, Aidan and Clifton, David A and Doherty, Aiden},
  journal={NPJ digital medicine},
  volume={7},
  number={1},
  pages={91},
  year={2024},
  publisher={Nature Publishing Group UK London},
  doi={10.1038/s41746-024-01062-3},
  note={doi: 10.1038/s41746-024-01062-3}
}

@inproceedings{that,
  title={Two-stream convolution augmented transformer for human activity recognition},
  author={Li, Bing and Cui, Wei and Wang, Wei and Zhang, Le and Chen, Zhenghua and Wu, Min},
  booktitle={Proceedings of the AAAI Conference on Artificial Intelligence},
  volume={35},
  number={1},
  pages={286--293},
  year={2021},
  doi={10.1609/aaai.v35i1.16103},
  note={doi: 10.1609/aaai.v35i1.16103}
}

@inproceedings{timesnet,
  title={{TimesNet}: Temporal {2D}-Variation Modeling for General Time Series Analysis},
  author={Wu, Haixu and Hu, Tengge and Liu, Yong and Zhou, Hang and Wang, Jianmin and Long, Mingsheng},
  booktitle={International Conference on Learning Representations},
  year={2023},
  url={https://mlanthology.org/iclr/2023/wu2023iclr-timesnet/}
}

@article{gpt4ts,
  title={One Fits All: Power General Time Series Analysis by Pretrained {LM}},
  author={Zhou, Tian and Niu, Peisong and Wang, Xue and Sun, Liang and Jin, Rong},
  journal={Advances in Neural Information Processing Systems},
  volume={36},
  pages={43322--43355},
  year={2023},
  doi={10.52202/075280-1877},
  note={doi: 10.52202/075280-1877}
}

@inproceedings{share,
  title={Unleashing the power of shared label structures for human activity recognition},
  author={Zhang, Xiyuan and Chowdhury, Ranak Roy and Zhang, Jiayun and Hong, Dezhi and Gupta, Rajesh K and Shang, Jingbo},
  booktitle={Proceedings of the 32nd ACM International Conference on Information and Knowledge Management},
  pages={3340--3350},
  year={2023},
  doi={10.1145/3583780.3615101},
  note={doi: 10.1145/3583780.3615101}
}

@inproceedings{raghar,
  title={{RAG-HAR}: Retrieval Augmented Generation-Based Human Activity Recognition},
  author={Sivaroopan, Nirhoshan and Karunarathna, Hansi and Madarasingha, Chamara and Jayasumana, Anura and Thilakarathna, Kanchana},
  booktitle={2026 IEEE International Conference on Pervasive Computing and Communications (PerCom)},
  pages={1--12},
  year={2026},
  doi={10.1109/PerCom67906.2026.11524560},
  note={doi: 10.1109/PerCom67906.2026.11524560}
}

@inproceedings{tent,
  author={Wang, Dequan and Shelhamer, Evan and Liu, Shaoteng and Olshausen, Bruno and Darrell, Trevor},
  title={{Tent}: Fully Test-Time Adaptation by Entropy Minimization},
  booktitle={International Conference on Learning Representations},
  year={2021},
  url={https://openreview.net/forum?id=uXl3bZLkr3c}
}

@article{cotta,
  title={Continual Test-Time Domain Adaptation},
  author={Qin Wang and Olga Fink and Luc Van Gool and Dengxin Dai},
  journal={2022 IEEE/CVF Conference on Computer Vision and Pattern Recognition (CVPR)},
  year={2022},
  pages={7191-7201},
  url={https://api.semanticscholar.org/CorpusID:247748613},
  doi={10.1109/CVPR52688.2022.00706},
  note={doi: 10.1109/CVPR52688.2022.00706}
}

@article{semanticloss,
  author={Arrotta, Luca and Civitarese, Gabriele and Bettini, Claudio},
  title={Semantic Loss: A New Neuro-Symbolic Approach for Context-Aware Human Activity Recognition},
  journal={Proc. ACM Interact. Mob. Wearable Ubiquitous Technol.},
  volume={7},
  number={4},
  year={2023},
  month=dec,
  issue_date={December 2023},
  online_publication_date={January 2024},
  articleno={147},
  numpages={29},
  doi={10.1145/3631407},
  note={Art. no. 147, doi: 10.1145/3631407}
}

\end{document}